# Combined Approach for Image Segmentation

Shradha Dakhare, Harshal Chowhan, Manoj B.Chandak

*Department of computer science & Engineering, Nagpur University*
*W.C.E.M. Dongargaon, R.C.O.E.M Nagpur, India*

*Abstract*— Many image segmentation techniques have been developed over the past two decades for segmenting the images, which help  for object recognition, occlusion boundary estimation within motion or stereo systems, image compression, image editing.
   In this, there is a combined approach for segmenting the image. By using histogram equalization to the input image, from which it gives contrast enhancement output image .After that by applying median filtering ,which will remove noise from contrast output image . At last I applied fuzzy c-mean clustering algorithm to denoising output image, which give segmented output image. In this way it produce better segmented image with less computation time.

*Keywords*— Histogram equalization, Median filter, Fuzzy C-Means.

## I. INTRODUCTION

Image segmentation refers to the major step in image Processing in which the inputs are images and, outputs are the attributes extracted from those images. The goal of segmentation is typically to locate certain objects of interest which may be depicted in the image. For example, in a vision guided car assembly system, the robot needs to pick up the appropriate components from the bin .For this segmentation followed by recognition is required. Its application area varies from the detection of cancerous cells to the identification of an airport from remote sensing data, etc. In all this area, the quality of final output depends largely on the quality of segmented output. Segmentation is the process of partitioning an image into non-intersecting regions such that each region is homogeneous and the union of no two adjacent regions is homogeneous. Formally, it can be defined as follows:

If F be the set of all pixels and P() be a uniformity (homogeneity)predicate defined on groups of connected pixels, then segmentation is a partitioning of the set F into a set of connected subsets or regions(S1, S2, · · · , Sn) such that

$$\bigcup_{i=1}^{n} S_i = F \text{ with } S_i \cap S_j = \text{null}, i \neq j$$

The uniformity predicate P(Si) = true for all regions (Si) and P(Si U Sj) = false when Si  is adjacent to Sj.

Segmentation divides image into its constituent Regions or objects. The level to which segmentation is carried out depends upon the problem being solved i.e. segmentation should stop when the objects of interest in an application have been isolated.
   Quantitative studies have been performed based on populations of biological Images**.** Such studies extremely require methods for segmentation, feature extraction, and classification. A first step in many analysis pipelines is segmentation, which can occur at several levels (e.g., separating nuclei, cells, tissues). This task has been an active field of research in image processing over the last 30 years, and various methods have been proposed and analyzed depending on the modality, quality, and resolution of the Microscopy images to analyze [2].

  This paper is organized as follows:  histogram equalization is used to increases the contrast enhancement of an image by increasing the dynamic range of intensity given to pixels with the most probable intensity values. Median Filter which is used for smoothing. Fuzzy C-means which is an *overlapping clustering* algorithm. In this paper we are combining these techniques for getting better quality segmented output.

## II. HISTOGRAM EQUALIZATION

   Histogram equalization is a kind of contrast enhancement that stretches the histogram so that all values occur (more or less) an equal number of times. This method usually increases the  global contrast of many images, especially when the usable data of the image is represented by close contrast values. Through this adjustment, the intensities can be better distributed on the histogram. This allows for areas of lower local contrast to gain a higher contrast. . The method is useful in images with backgrounds and foregrounds that are both bright or both dark. In particular, the method can lead to better views of bone structure in x-ray images, and to better detail in photographs that are over or under-exposed. A key advantage of the method is that it is a fairly straightforward technique and an invertible operator. The transformation is scaled such that the least intense value in the original image is mapped to a zero intensity value in the equalized image. As well, the most intense value in the original image is mapped to an intensity value that is equal to the maximum intensity value determined by the bit depth of the image. This produces results that have a dynamic range that is slightly larger than produced by the histogram equalization algorithm described in Gonzalez and Woods (2008). The algorithm has been tested to verify that performing further equalization on an already equalized image produces an output that is identical to the input. As well, it has been tested to verify that the histogram equalization of an input image with a constant probability distribution function produces an output image that is identical to the input. The Mat Lab script used to run these tests is provided in the Tests.mfile [3] original image on the left and the equalized image on the right. The histograms of the two images are shown below

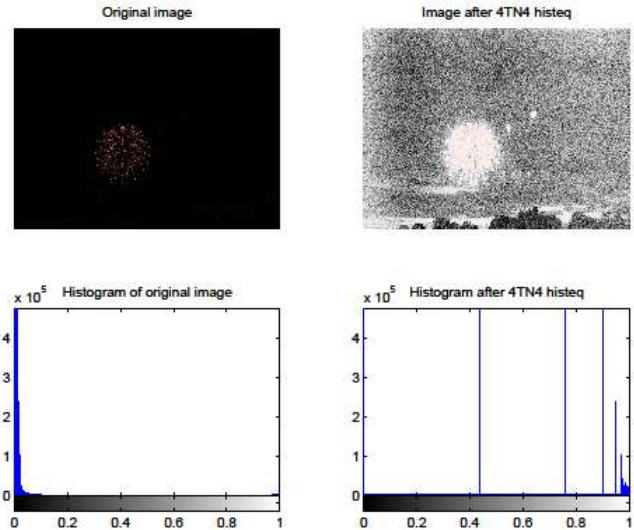

**Fig. 1 Histograms of two images**.

Fig. 1 this is an image taken at night at a fireworks display. Most pixels have a very low intensity. When the contrast of the dark pixels is increased using histogram equalization, the noise become very noticeable. However, this does a good job of bringing out details in the background. The trees and clouds near the bottom of the image are not visible in the original image.

A. Histogram Equalization Method

Consider a discrete grayscale image $\{x\}$ and let $n_i$ be the number of occurrences of gray level $i$. The probability of an occurrence of a pixel of level $i$ in the image is

$$p_x(i) = p(x = i) = \frac{n_i}{n}, \quad 0 \leq i < L$$

$L$ being the total number of gray levels in the image, $n$ being the total number of pixels in the image, and $p_x(i)$ being in fact the image. Histogram for pixel value i, normalized to [0,1].

The *cumulative distribution function* corresponding to $p_x$ as

$$cdf_x(i) = \sum_{j=0}^{i} p_x(j)$$

Which is also the image's accumulated normalized histogram? We would like to create a transformation of the form $y = T(x)$ to produce a new image $\{y\}$, such that its CDF will be linearized across the value range, i.e.

$$cdf_y(i) = iK$$

For some constant $K$. The properties of the CDF allow us to perform such a transform it is defined as

$$y = T(x) = cdf_x(x)$$

Notice that the T maps the levels into the range [0, 1]. In order to map the values back into their original range, the following simple transformation needs to be applied on the result:

$$y' = y \cdot (\max\{x\} - \min\{x\}) + \min\{x\}$$

The above describes histogram equalization on a grayscale image [4]. However it can also be used on color images by applying the same method separately to the Red, Green and Blue components of the RGB color values of the image. However, applying the same method on the Red, Green, and Blue components of an RGB image may yield dramatic changes in the image's color balance since the relative distributions of the color channels change as a result of applying the algorithm. However, if the image is first converted to another color space, Lab color space, or HSL/HSV color space in particular, then the algorithm can be applied to the luminance or value channel without resulting in changes to the hue and saturation of the image [5]. There are several histogram equalization methods in 3D space. Trapani's and Venetsanopoulos applied histogram equalization in 3D color space [6].

However, it results in "whitening" where the probability of bright pixels is higher than that of dark ones [7]. Han et al. proposed to use a new cdf defined by the is-luminance plane, which results in uniform gray distribution [8]

### III. MEDIAN FILTER

The median filter is a nonlinear digital filtering technique, often used to remove noise. To reduce noise several non-linear filters can be employed. One of the simplest techniques, the median filter, provided good noise reduction without affecting the borders of the objects on the image. The main idea of the median filter is to run through the signal entry by entry, replacing each entry with the median of neighboring entries. The pattern of neighbors is called the "window", which slides, entry by entry, over the entire signal. In the median filtering operation, the pixel values in the neighborhood window are ranked according to intensity, and the middle value (the median) becomes the output value for the pixel under evaluation.

| 123 | 125 | 126 | 130 | 140 |
|-----|-----|-----|-----|-----|
| 122 | 124 | 126 | 127 | 135 |
| 118 | 120 | 150 | 125 | 134 |
| 119 | 115 | 119 | 123 | 133 |
| 111 | 116 | 110 | 120 | 130 |

Neighborhood values:
115, 119, 120, 123,
124, 125, 127, 150

Median Values : 124

Fig. 2 calculating the median value of a pixel neighborhood. As can be seen, the central pixel value of 150 is rather unrepresentative of the surrounding pixels and is replaced with the median value: 124. A 3×3 square neighborhood is used here --- larger neighborhoods will produce more severe smoothing.

Often though, at the same time as reducing the noise in a signal, it is important to preserve the edges. Edges are of critical importance to the visual appearance of images, for example. For small to moderate levels of (Gaussian) noise, the median filter is demonstrably better than Gaussian blur at removing noise whilst preserving edges for a given, fixed window size[9]. However, its performance is not that much better than Gaussian blur for high levels of noise, whereas, for speckle noise12(impulsive noise), it is particularly effective [10]. Because of this, median filtering is very widely used in digital image processing.

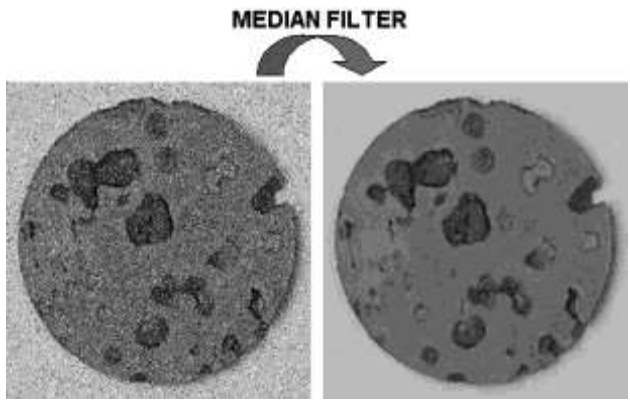

**Fig. 3 Median filters to improve an image corrupted by defective pixels**

## IV. FUZZY C-MEANS CLUSTERING

The fuzzy segmentation methods, which can retain more information from the original than hard segmentation methods [11-12].

In hard clustering, data is divided into distinct clusters, where each data element belongs to exactly one cluster. In fuzzy clustering (also referred to as soft clustering), data elements can belong to more than one cluster, and associated with each element is a set of membership levels. These indicate the strength of the association between that data element and a particular cluster. Fuzzy clustering is a process of assigning these membership levels, and then using them to assign data elements to one or more clusters [13].

One of the most widely used fuzzy clustering algorithms is the Fuzzy C-Means (FCM) Algorithm (Bezdek 1981). The FCM algorithm attempts to partition a finite collection of n elements $X = \{x_1,...,x_n\}$ into a collection of c fuzzy clusters with respect to some given criterion. Given a finite set of data, the algorithm returns a list of c cluster centers $C=\{c_1,...,c_c\}$ and a partition.

$J=U = u_{i,j} \in [0,1], i = 1,....,n, j = 1,...,c$

Where each element $u_{ij}$ tells the degree to which element $x_i$ belongs to cluster $c_j$.

In **fuzzy clustering**, each point has a degree of belonging to clusters, as in fuzzy logic, rather than belonging completely to just one cluster. Thus, points on the edge of a cluster may be *in the cluster* to a lesser degree than points in the center of cluster.

Any point $x$ has a set of coefficients giving the degree of being in the $k$th cluster $w_k(x)$. With fuzzy $c$-means, the centroid of a cluster is the mean of all points, weighted by their degree of belonging to the cluster:

$$C_k = \Sigma_x w_k(x) x / \Sigma_x w_k(x)$$

The degree of belonging, $w_k(x)$, is related inversely to the distance from $x$ to the cluster center as calculated on the previous pass. It also depends on a parameter $m$ that controls how much weight is given to the closest centre[14].

A. The Algorithm of Fuzzy C-means Clustering

- Step1. Choose a number of clusters in a given image.
- Step2. Assign randomly to each point coefficients for being in a cluster.
- Step3. Repeat until convergence criterion is met.
- Step4. Compute the center of each cluster.
- Step5. For each point, compute its coefficients of being in the cluster[4-5].

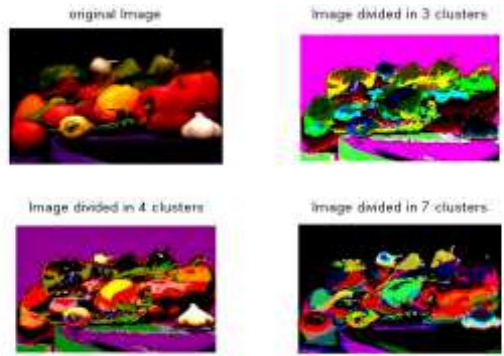

**Fig .4 Shown image segmentation by Fuzzy c means**

## V. WORKING OF COMBINED APPROACH FOR IMAGE SEGMENTATION

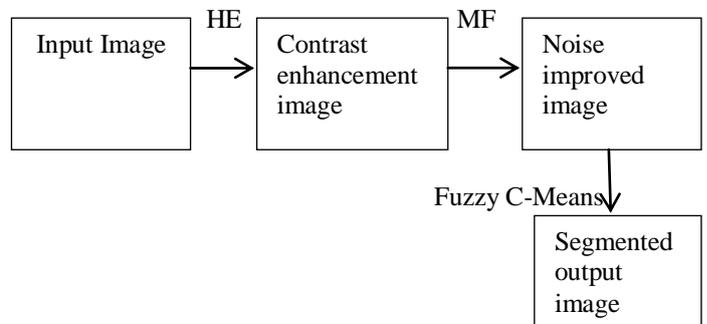

**Fig .5 combined approach used for segmentation**

The above fig.5 shows the combined approach which is used for segmentation. The histogram equalization is applied to the input image, from which it give contrast enhancement output image .After that by applying median filter ,which will remove noise from contrast output image .At last by applying fuzzy c-mean clustering algorithm to denoising output image, it produces higher accuracy segmented output image, with less computation time.

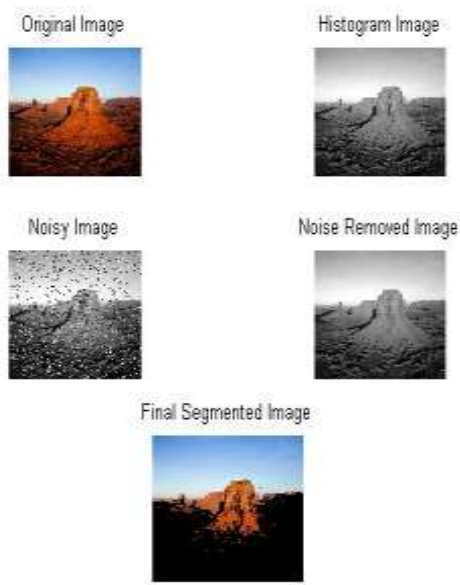

**Fig. 6 Show image segmentation by using combined approach**

In above figure an Desert Image is given ,by applying Histogram Equalization ,the contrast of image is increased. The noise which is present in this is removed by using Median Filter .The final segmented Image is obtained by using Fuzzy C Means.

## VI. CONCLUSIONS

The techniques are developed in MATLAB lab for analysis and comparisons. By combined approach that is by using Histogram Equalization, Median Filter and FCM , it produces higher accuracy and require less computation time.